\documentclass[10pt,twocolumn,letterpaper]{article}

\usepackage[pagenumbers]{cvpr} 
\usepackage[normalem]{ulem}

\usepackage{multirow}

\definecolor{cvprblue}{rgb}{0.21,0.49,0.74}
\usepackage[pagebackref,breaklinks,colorlinks,allcolors=cvprblue]{hyperref}

\def\paperID{ } 
\def\confName{CVPR}
\def\confYear{2026}

\title{VGGT-Segmentor: Geometry-Enhanced Cross-View Segmentation}

\author{Yulu Gao\footnotemark[1]\\
Hangzhou International Innovation\\
Institute of Beihang University\\
 Hangzhou, China\\
{\tt\small gyl97@buaa.edu.cn}
\and
Bohao Zhang\footnotemark[1]\\
Beihang University\\
Beijing, China\\
{\tt\small zbbhhh@buaa.edu.cn}
\and
Zongheng Tang\\
Hangzhou International Innovation\\
Institute of Beihang University\\
 Hangzhou, China\\
{\tt\small tzhhhh123@buaa.edu.cn}
\and
Jitong Liao\\
Beihang University\\
 Beijing, China\\
{\tt\small jitongliao@buaa.edu.cn}
\and
Wenjun Wu\\
Beihang University\\
Beijing, China\\
{\tt\small wwj09315@buaa.edu.cn}
\and
Si Liu\footnotemark[2]\\
Beihang University\\
Beijing, China\\
{\tt\small liusi@buaa.edu.cn}
}

\begin{document}
\maketitle
\renewcommand{\thefootnote}{\fnsymbol{footnote}}
\footnotetext[1]{These authors contributed equally.}
\footnotetext[2]{Corresponding author.}

\begin{abstract}
Instance-level object segmentation across disparate egocentric and exocentric views is a fundamental challenge in visual understanding, critical for applications in embodied AI and remote collaboration. This task is difficult due to severe changes in scale, perspective, and occlusion, which destabilize direct pixel-level matching. While recent geometry-aware models like VGGT provide a strong foundation for feature alignment, we find they often fail at dense prediction tasks due to significant pixel-level projection drift, even when their internal object-level attention remains consistent. To bridge this gap, we introduce VGGT-Segmentor (VGGT-S), a framework that unifies robust geometric modeling with pixel-accurate semantic segmentation. VGGT-S leverages VGGT's powerful cross-view feature representation and introduces a novel Union Segmentation Head. This head operates in three stages: mask prompt fusion, point-guided prediction, and iterative mask refinement, effectively translating high-level feature alignment into a precise segmentation mask. Furthermore, we propose a single-image self-supervised training strategy that eliminates the need for paired annotations and enables strong generalization. On the Ego–Exo4D benchmark, VGGT-S sets a new SOTA, achieving 67.7\% and 68.0\% average IoU for Ego→Exo and Exo→Ego tasks, respectively, significantly outperforming prior methods. Notably, our correspondence-free pretrained model surpasses most fully-supervised baselines, demonstrating the effectiveness and scalability of our approach. Code is publicly available at: \url{https://github.com/buaa-colalab/VGGT-S}.
\end{abstract}

\vspace{-2mm}
\section{Introduction}

Achieving instance-level correspondence across vastly different viewpoints is a key challenge in multi-view visual understanding, driving applications in embodied AI~\cite{kareer2025egomimic,eze2025cvmanip} and remote collaboration systems~\cite{bayro2025object, he2026bridging}. While traditional multi-view methods such as multi-view stereo~\cite{seitz1999photorealistic,furukawa2015multi,huang2018deepmvs} have significantly advanced scene geometry and keypoint correspondence, instance-level cross-view semantic correspondence, which concerns finding and segmenting the same physical object in two separate views, remains a largely underexplored frontier.

With the release of the large-scale Ego–Exo4D dataset~\cite{grauman2024ego}, researchers can now systematically investigate the ego–exo object correspondence task. Given an object mask as a query in one view, the goal is to locate and segment the same physical entity in another view. This capability is crucial for embodied intelligence and remote collaboration systems, as it enables the observation of key manipulated objects from an external viewpoint and provides real-time guidance or prompts in the first-person view.

The task is highly challenging due to the significant differences in scale, perspective, and occlusion between the two views. The ego camera is positioned close to the operator's hands, while the exo camera is often farther away or at a different height, causing the same object to appear differently in each view. Ego frames are frequently occluded by hands and tools, whereas exo frames contain numerous distractor objects and complex backgrounds, making pixel-level matching unstable.

\begin{figure*}[!htbp]        
  \centering
  \includegraphics[width=\textwidth]{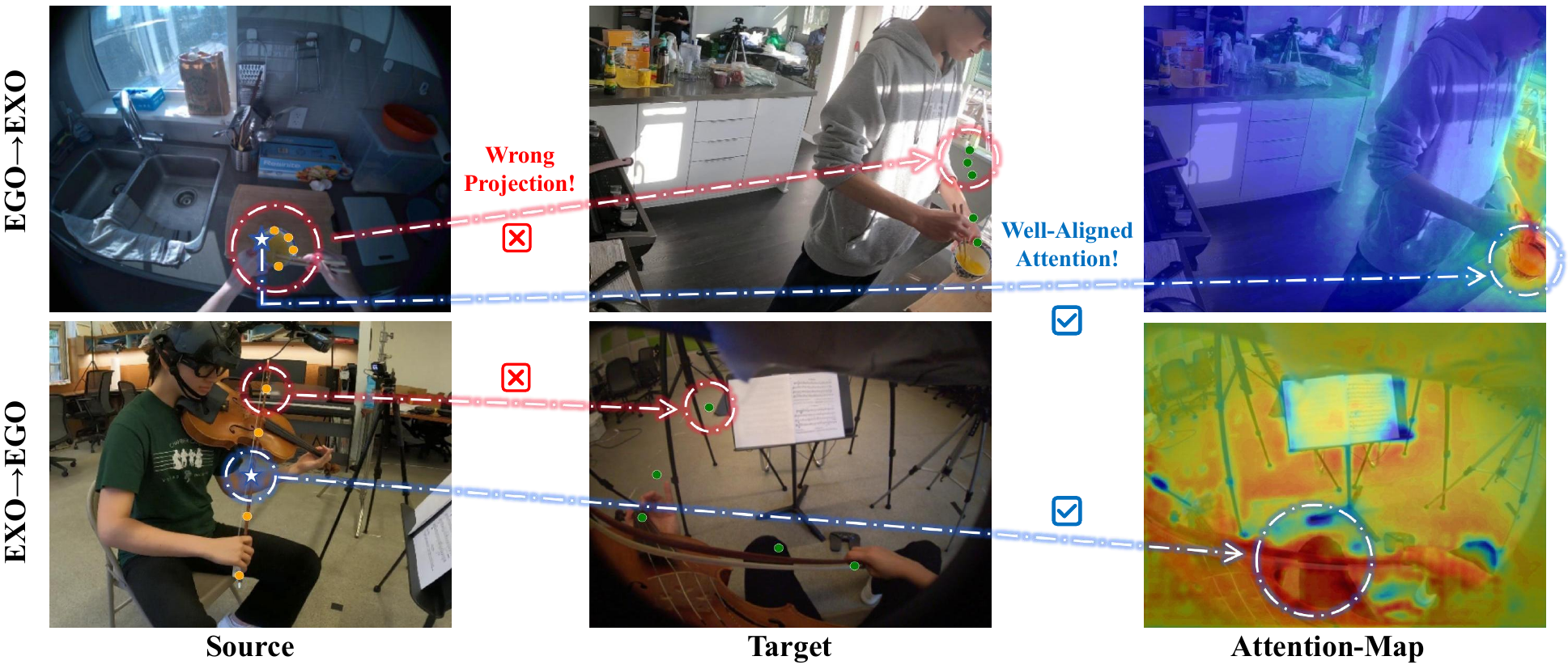}
\caption{\textbf{Visualizing VGGT Cross-View Correspondence.} Left: source image. Middle: target image with the projections of source-sampled points obtained by directly applying VGGT, which exhibit the systematic drift and misalignment. Right: star markers in the source image with the corresponding attention map on the target image, illustrating VGGT’s instance-consistent object alignment across views.}
  \label{fig:fig1}
\end{figure*}

Early works often rely on semantic consistency~\cite{liao2025domr} or the contextual understanding provided by large language models~\cite{zhang2024psalm,fu2025objectrelator}, but they tend to overlook geometric structures and spatial relationships. VGGT~\cite{wang2025vggt} offers a novel perspective. As a large transformer driven by visual geometry, VGGT jointly infers scene depth, camera parameters, and point maps across multiple views, enabling consistent modeling of both geometry and appearance. This provides a robust foundation for cross-view feature alignment. 

However, our study reveals a critical challenge in applying VGGT directly to dense segmentation: in ego–exo scenarios, severe occlusion and large viewpoint changes can cause its pixel-level point projections to drift, as illustrated in Figure~\ref{fig:fig1}. Notably, while the raw point tracking shows instability, VGGT's internal feature alignment remains consistently reliable, successfully focusing on the approximate object region.

To this end, we propose VGGT-Segmentor~(VGGT-S).  The model leverages VGGT’s strengths in cross-view feature modeling and introduces an object-level union segmentation head, which integrates the object mask as an explicit query into the cross-view reasoning process. The pipeline consists of three stages. 
The first is Mask Prompt Fusion, where two-view images are encoded by VGGT and then fused with the source-view object mask feature. This is followed by Point-Guided Prediction, where VGGT tracks points from the source mask and outputs a set of coarsely projected points in the target view to guide the fused features. The final stage is Mask Refinement, 
which refines the predicted mask by iteratively optimizing object boundaries and filling occluded regions.
Additionally, we propose a Single-Image Self-Supervised Training strategy that enables training without costly paired annotations, leading to powerful generalization.

On the Ego–Exo4D benchmark, VGGT-S achieves state-of-the-art average IoU scores of 67.7\% (Ego→Exo) and 68.0\% (Exo→Ego), outperforming the previous best methods by 18.0\% and 12.8\%, respectively.
Remarkably, even our correspondence-free pretrained VGGT-S variant surpasses prior fully-supervised baselines, highlighting its potential for scalable cross-view understanding without paired annotations.

Our key contributions are as follows:
\begin{itemize}
\item We introduce VGGT-S, a geometry-enhanced cross-view segmentation framework that fully exploits VGGT’s multi-view geometric representations.
\item We design the Union Segmentation Head, which comprises three coordinated stages including Mask Prompt Fusion, Point-Guided Prediction, and Mask Refinement, enabling robust cross-view segmentation.
\item We propose a Single-Image Self-Supervised Training strategy that reduces the need for paired annotations while enabling superior generalization for both Ego$\to$Exo and Exo$\to$Ego cross-view segmentation.
\item We achieve state-of-the-art results on the Ego–Exo4D benchmark, significantly surpassing previous methods.
\end{itemize}

\section{Related Work}
\label{sec:related}

\subsection{Cross-View Modeling}

Cross-view alignment and multi-view modeling are key directions in 3D vision. 
Classical structure-from-motion~\cite{lowe2004distinctive, calonder2010brief, agarwal2011building, schonberger2016structure, yi2016lift, wei2020deepsfm, shi2022clustergnn, wang2024vggsfm} and multi-view stereo methods~\cite{furukawa2015multi, galliani2015massively, fu2022geo, niemeyer2020differentiable, peng2022rethinking, ma2022multiview, zhang2023geomvsnet} rely on keypoint matching and geometric constraints to accurately reconstruct camera parameters and dense geometry in static scenes, but they are computationally demanding and struggle with non-rigid motion and large baselines. 
End-to-end neural methods have gradually reduced the need for traditional geometric optimization. VGGT~\cite{wang2025vggt} employs a large transformer in a feed-forward manner to jointly predict camera parameters, depth, and point maps, delivering efficient and accurate reconstruction without complex post-processing and serving as a geometry-consistent backbone for downstream tasks. Methods such as DUSt3R~\cite{wang2024dust3r} and MASt3R~\cite{leroy2024grounding} are related but often still depend on post-optimization. Given the substantial viewpoint differences in the ego–exo setting, pure reconstruction or two-view matching does not transfer directly to instance-level correspondence, motivating a unified approach that combines geometric structure and contextual semantics for instance-level correspondence. SegMASt3R~\cite{jayanti2025segmast3r} is a successful example of cross-view object segmentation that leverages 3D geometric priors to establish correspondences.

\subsection{Visual Object Correspondence}

Instance-level correspondence aims to establish matches for object instances across different views~\cite{grauman2024ego}. Some previous studies work on cross-view person matching~\cite{ardeshir2016ego2top, xu2018joint, fan2017identifying, wen2021seeing}. In the ego–exo setting, this problem is referred to as object correspondence. 
XSegTx~\cite{grauman2024ego} adapts a cross-image transformer architecture, conditioning on a query mask to perform mutual attention between egocentric and exocentric frames for joint mask prediction. XView-XMem~\cite{grauman2024ego} enhances tracking across interleaved ego-exo sequences by integrating embeddings from XSegTx into a working memory module to mitigate track drift.
PSALM~\cite{zhang2024psalm} combines a segmentation model with a large language model to tackle this task in a zero-shot manner. 
ObjectRelator~\cite{fu2025objectrelator} enhances PSALM by fusing language descriptions with visual queries and explicitly aligning object representations across different views to improve consistency. 
DOMR~\cite{liao2025domr} proposes a Dense Object Matching framework that pairs objects across views by jointly modeling visual, spatial, and semantic cues, 
modeling the contextual relationships among multiple objects simultaneously to suppress ambiguous matches.

\subsection{Segmentation Models}

Segmentation is fundamental to visual understanding, including semantic segmentation~\cite{chen2014semantic,chen2017rethinking,chen2017deeplab,chen2018encoder}, instance segmentation~\cite{hafiz2020survey,liu2018path,bolya2019yolact}, and panoptic segmentation~\cite{kirillov2019panoptic}. 
Recent unified segmentation models like Mask2Former~\cite{cheng2022masked}, along with multimodal promptable approaches such as SEEM~\cite{zou2023segment} and large-scale promptable models like SAM~\cite{kirillov2023segment} and SAM 2~\cite{ravi2024sam}, have demonstrated strong generalization on large datasets. However, most existing segmentation methods are single-view and lack cross-view alignment mechanisms.
MASA~\cite{li2024matching} leverages SAM's rich segmentation outputs to establish instance-level correspondences through extensive data transformations. Its core innovation lies in a self-training strategy that bootstraps instance associations from unlabeled images by applying geometric transformations to create pixel-level correspondences. These are then lifted by SAM to the instance level for contrastive similarity learning, enabling robust zero-shot tracking.

\begin{figure*}[!htbp]       
  \centering
  \includegraphics[width=\textwidth]{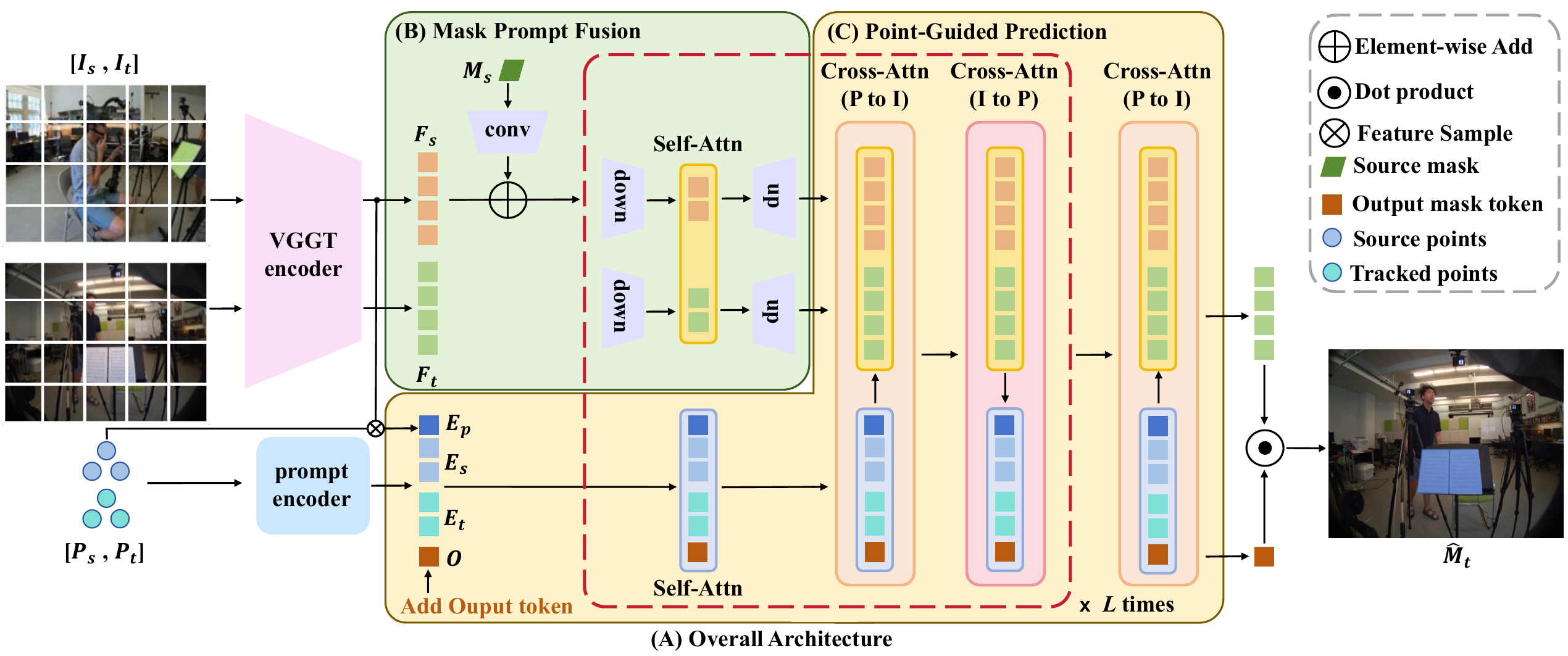}
\caption{%
\textbf{(A)} \textbf{Overall Architecture of VGGT-S}, which integrates the original VGGT encoder with our \textbf{Union Segmentation Head}. 
\textbf{(B) {Mask Prompt Fusion}} stage, which injects the source mask $M_s$ into source feature map $F_s$ and target feature map $F_t$ via convolutional fusion and a  \textbf{Bottleneck Fusion} module.
\textbf{(C) {Point-Guided Prediction}} stage, which uses point sets $(P_s, P_t)$ to guide target mask prediction through bidirectional interactions between point embeddings and image features.%
}
  \label{fig:method}
\end{figure*}

\section{Method}

\subsection{Overview}
VGGT \cite{wang2025vggt} is a vision model for multi-view geometric consistency, using a unified encoder with integrated tracking and feature interaction to model dense features. As illustrated in Figure~\ref{fig:method}(A), VGGT-S augments the VGGT encoder with a lightweight \textbf{Union Segmentation Head} that converts cross-view geometric cues into target-view masks. 
Given a source–target image pair $(I_{s}, I_{t})$ (e.g., Exo$\rightarrow$Ego), the VGGT encoder produces dense feature maps $F_{s}$ and $F_{t}$. 
The source mask $M_{s}$ is encoded and integrated into cross-view feature interactions. 
A compact set of representative points sampled from $M_{s}$ is tracked to the target frame via the VGGT's track head, generating $P_{t}$. 
These point prompts guide the prediction of the target mask $\hat{M}_{t}$ on $F_{t}$. 
During training, the VGGT encoder remains frozen and only the Union Segmentation Head is optimized, keeping the framework end-to-end while minimizing computational and memory overhead.

\subsection{VGGT Encoder}
Following VGGT, each image is patchified by a DINO-style~\cite{caron2021emerging} stem, which refers to a ViT-based patch embedding approach that splits images into patches and embeds them as tokens. They are then processed together through alternating frame-wise and global self-attention layers. A DPT-style~\cite{ranftl2021vision} head, which is a decoder for dense prediction that upsamples and fuses tokens into spatial feature maps, transforms tokens into dense feature maps geometrically aligned with depth, point, and tracking information. 
We extract these maps as inputs to our head:
\begin{equation}
x_{s}=\mathrm{Stem}(I_{s}), \quad x_{t}=\mathrm{Stem}(I_{t}),
\end{equation}
\begin{equation}
h_{s}, h_{t}=\mathrm{VGGT}(x_{s}, x_{t}),
\end{equation}
\begin{equation}
F_{s}, F_{t}=\mathrm{DPT}(h_{s}, h_{t}).
\end{equation}
The resulting geometry-aware features $F_{s}$ and $F_{t}$ are fed into the Union Segmentation Head.

\subsection{Union Segmentation Head}
The Union Segmentation Head consists of three stages, Mask Prompt Fusion, Point-Guided Prediction and Mask Refinement.

\noindent\textbf{Mask Prompt Fusion.}
As shown in Figure~\ref{fig:method}(B), we first encode the source mask $M_{s}$  into a high-dimensional embedding that captures its spatial layout and identity:
\begin{equation}
E_{m}=\mathrm{Conv}(M_{s}).
\end{equation}
This embedding $E_{m}$ is added to the source features $F_{s}$ directly:
\begin{equation}
F_{s}^{\prime}=F_{s}+ E_{m}.
\end{equation}
Although $M_{s}$ is now fused into $F_{s}^{\prime}$, it has not yet interacted sufficiently with $F_{t}$. 
Therefore, we introduce a \textbf{Bottleneck Fusion} module that integrates self-attention (SelfAttn), feed-forward network (FFN) as well as downsampling $\mathrm{D}_{r}$ and upsampling $\mathrm{U}_{r}$ (ratio $r$):
\begin{equation}
\tilde{F}_{s}=\mathrm{D}_{r}(F_{s}^{\prime}), \quad \tilde{F}_{t}=\mathrm{D}_{r}(F_{t}),
\end{equation}
\begin{equation}
\dot{F}_{s},\,\dot{F}_{t}=\mathrm{FFN\big(}\mathrm{SelfAttn}\big([\tilde{F}_{s}\ ,\ \tilde{F}_{t}]\big)\big),
\end{equation}
\begin{equation}
F_{s}^{\star}=\mathrm{U}_{r}(\dot{F}_{s}), \quad F_{t}^{\star}=\mathrm{U}_{r}(\dot{F}_{t}).
\end{equation}
Here $[\cdot\ ,\ \cdot]$ denotes concatenation. 
The resulting $F^{\star}=[F_{s}^{\star}, F_{t}^{\star}]$ is a compact yet expressive representation containing both geometric and semantic cues from two views.

\noindent\textbf{Point-Guided Prediction.}
We next generate point prompts from the source mask. 
Let the foreground pixel set be
\begin{equation}
\Omega=\{(x,y)\ |\ M_{s}(x,y)=1\}.
\end{equation}
We sample $K_{\text{pt}}$ representative points using K-Means algorithm~\cite{lloyd1982least}:
\begin{equation}
P_{s}=\mathrm{kmeans}(\Omega,K_{\text{pt}}).
\end{equation}
VGGT's track head $\mathcal{T}$ projects them to the target frame:
\begin{equation}
P_{t}=\mathcal{T}(P_{s}; I_{s}, I_{t}).
\end{equation}
A prompt encoder $\psi$ maps points to embeddings, and a learnable output mask token $O$ together with source point features sampled from $F_s$ are appended:
\begin{equation}
    E_{p}=\mathcal{G}(P_{s}, F_s), \quad E_{s}=\psi(P_{s}), \quad E_{t}=\psi(P_{t}), 
\end{equation}
\begin{equation}
\quad Q_{0}=[E_{p}, E_{s},E_{t},O],
\end{equation}
where $Q_{0}$ denotes the prompt queries.

As shown in Figure~\ref{fig:method}(C), we apply $L$-layer lightweight decoder blocks, each consisting of self-attention among prompts, followed by point-to-image and image-to-point cross-attention (CrossAttn):
\begin{equation}
\bar{Q}_{\ell}=\mathrm{SelfAttn}(Q_{\ell-1}),
\end{equation}
\begin{equation}
Q_{\ell}=\mathrm{CrossAttn}_{P\rightarrow I}(\bar{Q}_{\ell},F_{\ell}^{\star}),
\end{equation}
\begin{equation}
H_{\ell}=\mathrm{CrossAttn}_{I\rightarrow P}(F_{\ell}^{\star},Q_{\ell}), \quad \ell=1,\ldots,L.
\end{equation}
where $F_{\ell}^{\star}$ denotes the output of the Bottleneck Fusion module within the $\ell$-th block, and $H_{\ell}$ represents the resulting fused image features produced by the same block.

Finally, we perform an additional point-to-image cross-attention using the refined output mask token $O_{L}$, and generate an initial mask through per-pixel dot products on $H_{t}$, which corresponds to the target-view component of the final fused image features $H_{L}$:
\begin{equation}
\tilde{O}=\mathrm{CrossAttn}_{P\rightarrow I}(O_{L},H_{t}),
\end{equation}
\begin{equation}
z(x,y)=\big(W\tilde{O}+b\big)^{\!\top}\mathbf{f}_{t}(x,y), 
\end{equation}
\begin{equation}
\quad \hat{M}_{t}^{(0)}(x,y)=\sigma\!\big(z(x,y)\big),
\end{equation}
where $W$ and $b$ denote the weights and bias of an MLP, $\mathbf{f}_{t}(x,y)$ is the feature vector at pixel position $(x,y)$ on $H_{t}$ and $\sigma(\cdot)$ is the sigmoid function.

\noindent\textbf{Mask Refinement.}
To sharpen boundaries and handle occlusions, we adopt an iterative refinement module. 
At iteration $k$,
\begin{equation}
\hat{M}_{t}^{(k+1)}=\Psi\big(F_{s},\,M_{s},\,F_{t},\,\hat{M}_{t}^{(k)},\,Q\big),
\end{equation}
where $\Psi$ denotes our lightweight mask decoder, $Q$ denotes the refined prompt queries.

 During training, we perform refinement iterations and backpropagate gradients only through the final iteration and half of the samples in each batch undergo refinement, while the other half do not. This process progressively sharpens object boundaries, fills occluded regions, and improves cross-view segmentation quality. More details are in the Supplementary Material.

\subsection{Single-Image Self-Supervised Training}
To reduce reliance on paired annotations and enhance generalization, we introduce a Single-Image Self-Supervised Training strategy inspired by the augmentation methods of MASA~\cite{li2024matching}. 
Given any image $I$, we generate an augmented view $I^{\prime}$ and obtain a pseudo mask $M$ from an offline segmentor~\cite{kirillov2023segment}. 
The model is required to predict the same object’s mask $\hat{M}^{\prime}$ on $I^{\prime}$.

The training strategy employs dynamic augmentations from two families: 
(1) VGGT-adaptive (e.g., scaling, mild rotations, cropping), which preserve VGGT’s point mapping. 
In this case, both views are processed through the VGGT encoder, and the VGGT’s track head provides point prompts on the target view. 
(2) VGGT-non-adaptive (e.g., large rotations, horizontal flips), which heavily disrupt cross-view alignment and cause VGGT to fail in maintaining effective correspondence.
Here, the two views are processed independently by VGGT encoder, and we perturb target ground-truth points to synthesize prompts. 
By mixing these two families, the model learns a cross-view mask head well aligned with VGGT features. It can recover target masks under substantial viewpoint changes, enabling robust Ego$\rightarrow$Exo and Exo$\rightarrow$Ego transfer without paired annotations.

Specifically, we train the model on a $1\,/\,20$ subset of the SA-1B dataset~\cite{ravi2024sam} to obtain a correspondence-free pretrained variant. When evaluated on the Ego-Exo4D dataset, this variant still delivers competitive results.

\section{Experiments}

\subsection{Setup and Implementation Details}

\noindent\textbf{Dataset.} We use the ego–exo correspondence benchmark from the Ego-Exo4D dataset~\cite{grauman2024ego}, which contains synchronized first-person and third-person videos of professional skill demonstrations across various domains. The dataset includes 1,335 annotated takes and 5,566 target objects. It provides 1.8 million masks sampled at 1 FPS, of which 742K are egocentric and 1.1 million are exocentric. On average, each video consists of approximately 5.5 objects and 173 frames per track. The annotations cover a wide range of objects, including tools, relevant environmental items, and human body parts. We use the official train/validation split for our experiments, and the evaluation metric is the mean Intersection over Union (IoU) between predicted and ground-truth masks.

\begin{table*}[!t]
    \centering
    \caption{Comparison with prior methods on Ego-Exo4D dataset. “ZSL” denotes the zero-shot learning results. “Type S” denotes spatial-only modeling, while “Type ST” denotes spatio-temporal modeling. Our VGGT-S provides both supervised and zero-shot learning results.}
    \label{tab:main-result}
    \begin{tabular}{lccc|lccc}
        \toprule
        \multicolumn{4}{c|}{\textbf{Ego$\to$Exo}} & \multicolumn{4}{c}{\textbf{Exo$\to$Ego}} \\
        \textbf{Method} & \textbf{ZSL} & \textbf{Type} & \textbf{IoU} $\uparrow$ & \textbf{Method} & \textbf{ZSL} & \textbf{Type} & \textbf{IoU} $\uparrow$ \\
        \midrule
        XSegTx~\cite{grauman2024ego} & $\checkmark$ & S & 0.3 & XSegTx~\cite{grauman2024ego} & $\checkmark$ & S & 1.3 \\
        SEEM~\cite{zou2023segment} & $\checkmark$ & S & 1.1 & SEEM~\cite{zou2023segment} & $\checkmark$ & S & 4.1 \\
        XSegTx~\cite{grauman2024ego} & $\times$ & S & 6.2 & XSegTx~\cite{grauman2024ego} & $\times$ & S & 30.2 \\
        CMX~\cite{zhang2023cmx} & $\times$ & S & 6.8 & CMX~\cite{zhang2023cmx} & $\times$ & S & 12.0 \\
        PSALM~\cite{zhang2024psalm} & $\checkmark$ & S & 7.9 & PSALM~\cite{zhang2024psalm} & $\checkmark$ & S & 9.6 \\
        DCAMA~\cite{shi2022dense} & $\checkmark$ & S & 9.7 & DCAMA~\cite{shi2022dense} & $\checkmark$ & S & 14.1 \\
        XView-XMem~\cite{grauman2024ego} & $\checkmark$ & ST & 16.2 & XView-XMem~\cite{grauman2024ego} & $\checkmark$ & ST & 13.5 \\
        XView-XMem~\cite{grauman2024ego} & $\times$ & ST & 17.7 & XView-XMem~\cite{grauman2024ego} & $\times$ & ST & 20.7 \\
        XView-XMem + XSegTx~\cite{grauman2024ego} & $\times$ & ST & 36.9 & XView-XMem + XSegTx~\cite{grauman2024ego} & $\times$ & ST & 36.1 \\
        PCC~\cite{baade2025sscc} & $\times$ & S & 37.7 & PCC~\cite{baade2025sscc} & $\times$ & S & 43.7 \\
        PSALM~\cite{zhang2024psalm} & $\times$ & S & 41.3 & PSALM~\cite{zhang2024psalm} & $\times$ & S & 47.3 \\
        ObjectRelator~\cite{fu2025objectrelator} & $\times$ & S & 45.4 & ObjectRelator~\cite{fu2025objectrelator} & $\times$ & S & 50.9 \\
        DOMR~\cite{liao2025domr} & $\times$ & S & 49.7 & DOMR~\cite{liao2025domr}  & $\times$ & S & 55.2 \\
        \midrule
        \textbf{VGGT-S (Ours)} & $\checkmark$ & S & \textbf{54.1} & \textbf{VGGT-S } & $\checkmark$ & S & \textbf{58.4} \\
        \textbf{VGGT-S (Ours)} & $\times$ & S & \textbf{67.7} & \textbf{VGGT-S } & $\times$ & S & \textbf{68.0} \\
        \bottomrule
    \end{tabular}
\end{table*}

\noindent\textbf{Implementation Details.} 
We adopt the official VGGT encoder settings, using an image patch size of 14. In the Mask Prompt Fusion stage, we downsample the source mask through a convolution layer, reducing its size to half of the original resolution. This ensures consistency with the feature map of the image output by VGGT. In the Point-Guided Prediction stage, we apply the K-Means algorithm~\cite{lloyd1982least}, setting the number of clusters to 5 to match the number of sampled points. Clustering is refined only once to save training time. 
Following SAM~\cite{kirillov2023segment}, we supervise the model’s predictions using a linear combination of focal and dice losses with a weight ratio of $20:1$. For optimization, we use AdamW~\cite{loshchilov2017decoupled}, with an initial learning rate of $5 \times 10^{-5}$ and a weight decay of $1 \times 10^{-4}$. The model is trained for 12 epochs, with the learning rate reduced by a factor of 0.1 after 8 and 11 epochs. To prevent gradient explosion, we clip the $L_2$ norm of all gradients to 1.0.  All experiments are conducted on 4$\times$NVIDIA RTX 4090 GPUs, with a batch size of 8 during training. 
For inference speed, we run 100 forward passes on a single image using a single GPU and report the average time. 
In the Ego$\to$Exo task, the remapping strategy introduces an additional mapping step, which is omitted in the subsequent time measurements. We also adopt a cropping strategy. Both are detailed in the Supplementary Material.

\subsection{Main Results}

We evaluate our method on the Ego-Exo4D benchmark and report the results in Table~\ref{tab:main-result}. Our approach achieves 67.7\% IoU on Ego$\to$Exo and 68.0\% IoU on Exo$\to$Ego, surpassing the previous state-of-the-art method, DOMR, by 18.0\% and 12.8\%, respectively. Compared to the LLM-based ObjectRelator, our method outperforms it by 22.3\% and 17.1\% in the two directions, while also demonstrating significantly higher efficiency during inference.

In the zero-shot setting, our model achieves 54.1\% IoU on Ego$\to$Exo and 58.4\% IoU on Exo$\to$Ego. We improve over PSALM by 46.2\% and 48.8\%, and over XView-XMem by 37.9\% and 44.9\%, respectively. Notably, XView-XMem leverages spatiotemporal cues, whereas our method relies solely on image-level features and still outperforms it. Our correspondence-free pretrained variant also surpasses the supervised method, DOMR, on both tasks, with gains of 4.4\% and 3.2\%, demonstrating strong generalization to unseen objects and scenes. 

To further validate the generalizability of VGGT-S, we finetune the correspondence-free pretrained model on the MvMHAT dataset~\cite{gan2021mvmhat} for 1 epoch. Surprisingly, the resulting AP reaches 80.7\%, surpassing DOMR by 9.6\% and the method in~\cite{gan2021mvmhat} by 16.9\%, as Table~\ref{tab:MvMHAT_dataset} shows.
These results demonstrate the strong generalization capability of our VGGT-S model.

\begin{table}[tb]
    \caption{Comparison with prior methods on MvMHAT dataset.}
    \label{tab:MvMHAT_dataset}
    \centering
    \begin{tabular}{cc}
        \toprule
         \textbf{Method} & \textbf{AP} \\
        \midrule
        MvMHAT~\cite{gan2021mvmhat}      & 63.8 \\
        DOMR~\cite{liao2025domr}   & 71.1  \\
        \midrule
        VGGT-S (Ours) & 80.7 \\
        \bottomrule
    \end{tabular}
\end{table}

\subsection{Ablation Studies}
\noindent\textbf{Component Analysis.}
A step-by-step ablation of the proposed components is provided in Table~\ref{tab:abl-component-analysis}. We begin with a Plain Head that encodes the source view mask and predicts the target mask using an output mask token, establishing a direct baseline. In the next step, adding Bottleneck Fusion leads to clear improvements, demonstrating that cross-view feature aggregation is crucial for viewpoint transfer, as target features gain spatial prior information from the source object. Introducing Point-Guided Prediction results in a significant increase in IoU by incorporating sparse, geometry-aware anchors, which are robust to perspective and scale changes. Finally, the Mask Refinement module consistently boosts IoU with minimal computational overhead by refining boundaries and correcting small misalignments. The full model, incorporating all components, achieves an overall improvement of 32.2\% on the Ego$\to$Exo task and 30.9\% on the Exo$\to$Ego task over the Plain Head setting, validating the effectiveness of the geometry-enhanced design.

\begin{table}[tb]
    \caption{Component analysis. ``BF'' denotes the Bottleneck Fusion module in Mask Prompt Fusion stage. ``PGP'' denotes the Point-Guided Prediction. ``MR'' denotes Mask Refinement stage.}
    \label{tab:abl-component-analysis}
    \centering
    \begin{tabular}{lccc}
        \toprule
        \multirow{2}{*}{\raisebox{-0.6ex}{\textbf{Method}}}  
        & \multicolumn{2}{c}{\textbf{IoU $\uparrow$}} 
        & \multirow{2}{*}{\raisebox{-0.6ex}{\textbf{Time (ms)}}} \\
        \cmidrule(lr){2-3}
          & \textbf{Ego$\to$Exo} & \textbf{Exo$\to$Ego} &  \\
        \midrule
        Plain Head      & 35.5 & 37.1 & 105.8 \\
        + BF    & 50.2 & 52.3 & 107.4 \\
        + PGP   & 62.2 & 63.5 & 153.2 \\
        + MR     & 67.7 & 68.0 & 161.4 \\
        \bottomrule
    \end{tabular}
\end{table}

\begin{table}[tb]
    \caption{Effect of Bottleneck Fusion resolution.}
    \label{tab:neck-size}
    \centering
    \begin{tabular}{cccc}
        \toprule
        \multirow{2}{*}{\raisebox{-0.6ex}{\textbf{Fusion Size}}} 
        & \multicolumn{2}{c}{\textbf{IoU $\uparrow$}} 
        & \multirow{2}{*}{\raisebox{-0.6ex}{\textbf{Time (ms)}}} \\
        \cmidrule(lr){2-3}
         & \textbf{Ego$\to$Exo} & \textbf{Exo$\to$Ego} & \\
        \midrule
        37×37   & 67.7 & 68.0 & 161.4 \\
        74×74   & 68.4 & 68.5 & 180.9 \\
        518×518 & $\textit{OOM}$ & $\textit{OOM}$ & $\textit{OOM}$ \\
        \bottomrule
    \end{tabular}
\end{table}

\begin{table}[tb]
    \caption{Effect of the number of points used in Point-Guided Prediction.}
    \label{tab:point-num}
    \centering
    \begin{tabular}{cccc}
        \toprule
        \multirow{2}{*}{\raisebox{-0.6ex}{\textbf{\#Points}}} 
        & \multicolumn{2}{c}{\textbf{IoU $\uparrow$}} 
        & \multirow{2}{*}{\raisebox{-0.6ex}{\textbf{Time (ms)}}} \\
        \cmidrule(lr){2-3}
         & \textbf{Ego$\to$Exo} & \textbf{Exo$\to$Ego} & \\
        \midrule
        1 & 61.5 & 63.4 & 160.1 \\
        5 & 67.7 & 68.0 & 161.4 \\
        9 & 68.3 & 68.5 & 162.9 \\
        \bottomrule
    \end{tabular}
\end{table}

\begin{table}[tb]
    \caption{Effect of iterations in Mask Refinement.}
    \label{tab:mask-refine-iters}
    \centering
    \begin{tabular}{cccc}
        \toprule
        \multirow{2}{*}{\raisebox{-0.6ex}{\textbf{\#Refine Iters}}} 
        & \multicolumn{2}{c}{\textbf{IoU $\uparrow$}} 
        & \multirow{2}{*}{\raisebox{-0.6ex}{\textbf{Time (ms)}}} \\
        \cmidrule(lr){2-3}
         & \textbf{Ego$\to$Exo} & \textbf{Exo$\to$Ego} & \\
        \midrule
        0 & 62.2 & 63.5 & 153.2 \\
        1 & 66.3 & 67.5 & 157.7 \\
        2 & 67.7 & 68.0 & 161.4 \\
        3 & 67.9 & 68.4 & 165.3 \\
        \bottomrule
    \end{tabular}
\end{table}

\begin{table}[tb]
    \caption{Effect of input image size.}
    \label{tab:image-size}
    \centering
    \begin{tabular}{cccc}
        \toprule
        \multirow{2}{*}{\raisebox{-0.6ex}{\textbf{Image Size}}} 
        & \multicolumn{2}{c}{\textbf{IoU $\uparrow$}} 
        & \multirow{2}{*}{\raisebox{-0.6ex}{\textbf{Time (ms)}}} \\
        \cmidrule(lr){2-3}
         & \textbf{Ego$\to$Exo} & \textbf{Exo$\to$Ego} & \\
        \midrule
        420×420 & 66.1 & 66.3 & 136.8 \\
        518×518 & 67.7 & 68.0 & 161.4 \\
        700×700 & 68.5 & 68.9 & 225.4 \\
        \bottomrule
    \end{tabular}
\end{table}

\begin{table}[tb]
    \caption{Effect of the number of decoder blocks.}
    \label{tab:fusion-block-num}
    \centering
    \begin{tabular}{cccc}
        \toprule
        \multirow{2}{*}{\raisebox{-0.6ex}{\textbf{\#Blocks}}} 
        & \multicolumn{2}{c}{\textbf{IoU $\uparrow$}} 
        & \multirow{2}{*}{\raisebox{-0.6ex}{\textbf{Time (ms)}}} \\
        \cmidrule(lr){2-3}
         & \textbf{Ego$\to$Exo} & \textbf{Exo$\to$Ego} & \\
        \midrule
        1 & 65.1 & 65.5 & 158.2 \\
        2 & 67.7 & 68.0 & 161.4 \\
        3 & 68.4 & 68.7 & 165.4 \\
        6 & 68.8 & 69.3 & 176.8 \\
        \bottomrule
    \end{tabular}
\end{table}

\noindent\textbf{Effect of Bottleneck Fusion Resolution.}
We investigate the impact of fusion resolution in the Bottleneck Fusion module at spatial sizes of 37×37, 74×74, and 518×518, as summarized in Table~\ref{tab:neck-size}. Increasing the resolution from 37×37 to 74×74 results in improvements of 0.7\% and 0.5\% IoU for the two tasks, respectively. However, this also increases latency due to the quadratic complexity of self-attention at higher spatial resolutions. Further scaling to 518×518 causes out-of-memory (\textit{OOM}) issues during training. Balancing both accuracy and efficiency, we adopt 37×37 as the default resolution for mask and image fusion in our main experiments, which retains most of the benefits of cross-view coupling while maintaining inference efficiency.

\noindent\textbf{Effect of the Number of Points.}
Table~\ref{tab:point-num} analyzes the impact of the number of points used in Point-Guided Prediction. Increasing the number of sampled points from 1 to 5 improves the IoU by 6.2\% and 4.6\% on the Ego$\to$Exo and Exo$\to$Ego tasks, respectively. Further increasing the number of points from 5 to 9 results in only marginal gains of 0.6\% and 0.5\% for the two tasks. We adopt 5 points for all final results. These experiments demonstrate that sparse points provide an effective and efficient guidance signal for cross-view segmentation.

\noindent\textbf{Effect of Mask Refinement Iterations.}
We vary the number of Mask Refinement iterations in Table~\ref{tab:mask-refine-iters}. As the number of iterations increases from 0 to 3, IoU improves from 62.2\% to 67.9\% on the Ego$\to$Exo task and from 63.5\% to 68.4\% on the Exo$\to$Ego task, resulting in total gains of +5.7\% and +4.9\%, respectively. Since each iteration re-invokes the mask head, the computational cost scales approximately linearly with the number of iterations. With our lightweight head, two iterations provide an optimal trade-off, delivering significant improvements over a single pass with minimal additional latency, while further iterations result in only marginal gains.

\begin{figure*}[!htbp]        
  \centering
  \includegraphics[width=0.95\textwidth]{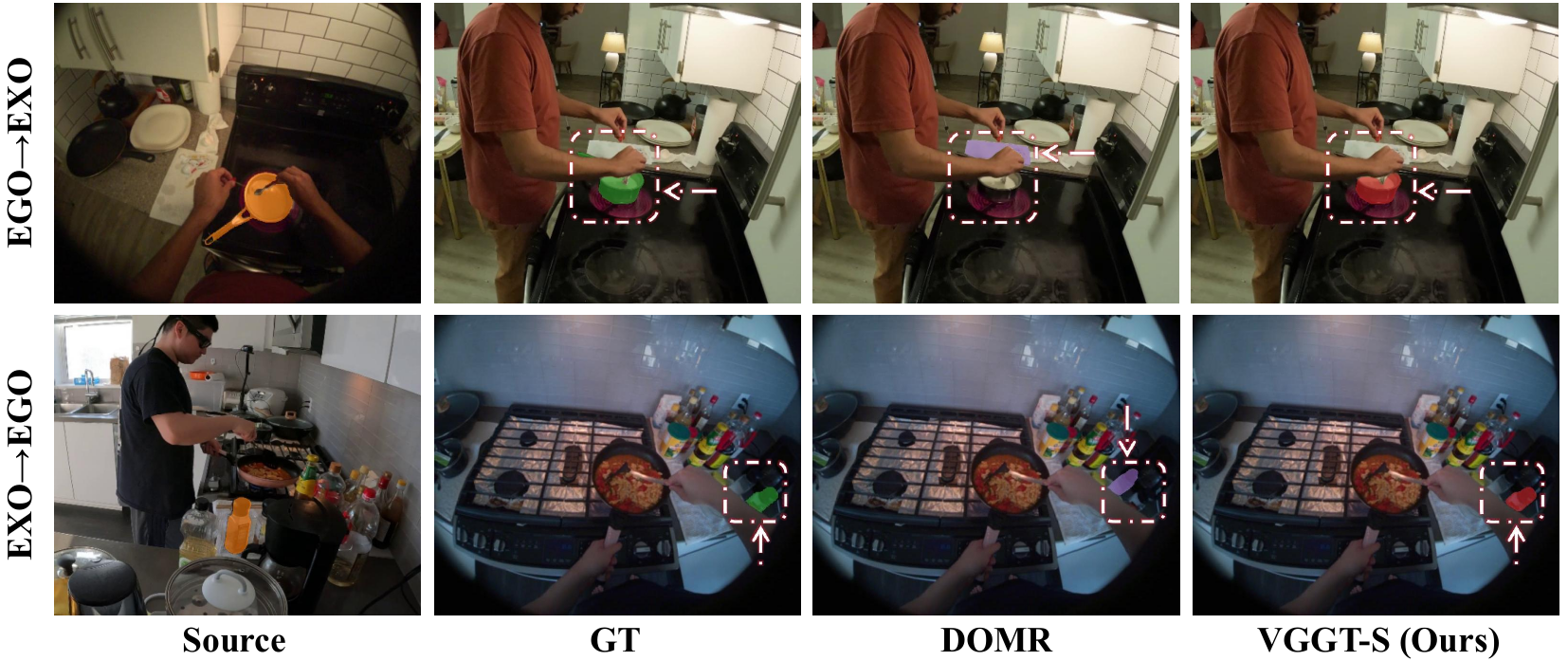}
\caption{\textbf{Visualization of VGGT-S vs. DOMR.} The first row shows the Ego$\to$Exo task. DOMR incorrectly takes the chopping board as the predicted result, while VGGT-S correctly identifies the pot. The second row illustrates the Exo$\to$Ego task. Two similar bottles are nearby. Due to a lack of geometric information, DOMR mistakenly confuses them, whereas VGGT-S continues to make accurate predictions.}
  \label{fig:Comparison}
\end{figure*}

\noindent\textbf{Effect of Input Image Size.} Table~\ref{tab:image-size} evaluates the impact of input resolutions 420×420, 518×518, and 700×700. While higher input resolutions lead to monotonic improvements in IoU, they also increase computational and memory requirements, resulting in higher latency and reduced throughput during inference. This trade-off is consistently observed across both Ego$\to$Exo and Exo$\to$Ego settings. Therefore, we adopt 518×518 as the default resolution, as it strikes a good balance between accuracy and efficiency for both directions, and aligns with our training time configuration and hardware profile.

\noindent\textbf{Effect of the Number of Decoder Blocks.}
Table~\ref{tab:fusion-block-num} ablates the number of decoder blocks. Performance improves steadily from 1 to 6 blocks, suggesting that deeper cross-view fusion enhances alignment and refines mask details. To maintain a compact and efficient model, we use 2 blocks by default in all reported results. This configuration captures most of the benefits from iterative point and image interactions without introducing noticeable slowdowns.

\subsection{Qualitative Results}

\noindent\textbf{Visualization of VGGT-S vs. DOMR.}
Figure~\ref{fig:Comparison} compares VGGT-S with DOMR on both Ego$\to$Exo and Exo$\to$Ego tasks. Leveraging geometry-enhanced cues, VGGT-S demonstrates clear advantages in spatial localization. Even under significant viewpoint changes and in the presence of visually similar distractors, our method effectively restricts the correspondence search to geometrically reasonable regions, ensuring consistent alignment between views. This geometric constraint reduces ambiguity during matching. As a result, VGGT-S more reliably identifies the correct target among multiple confusing proposals, producing cleaner and better-aligned masks with sharper boundaries, whereas DOMR tends to drift towards nearby look-alike objects, exhibits unstable correspondences, and often leads to noticeable boundary misalignment.

\begin{figure}[!htbp]
    \centering
    \includegraphics[width=\linewidth]{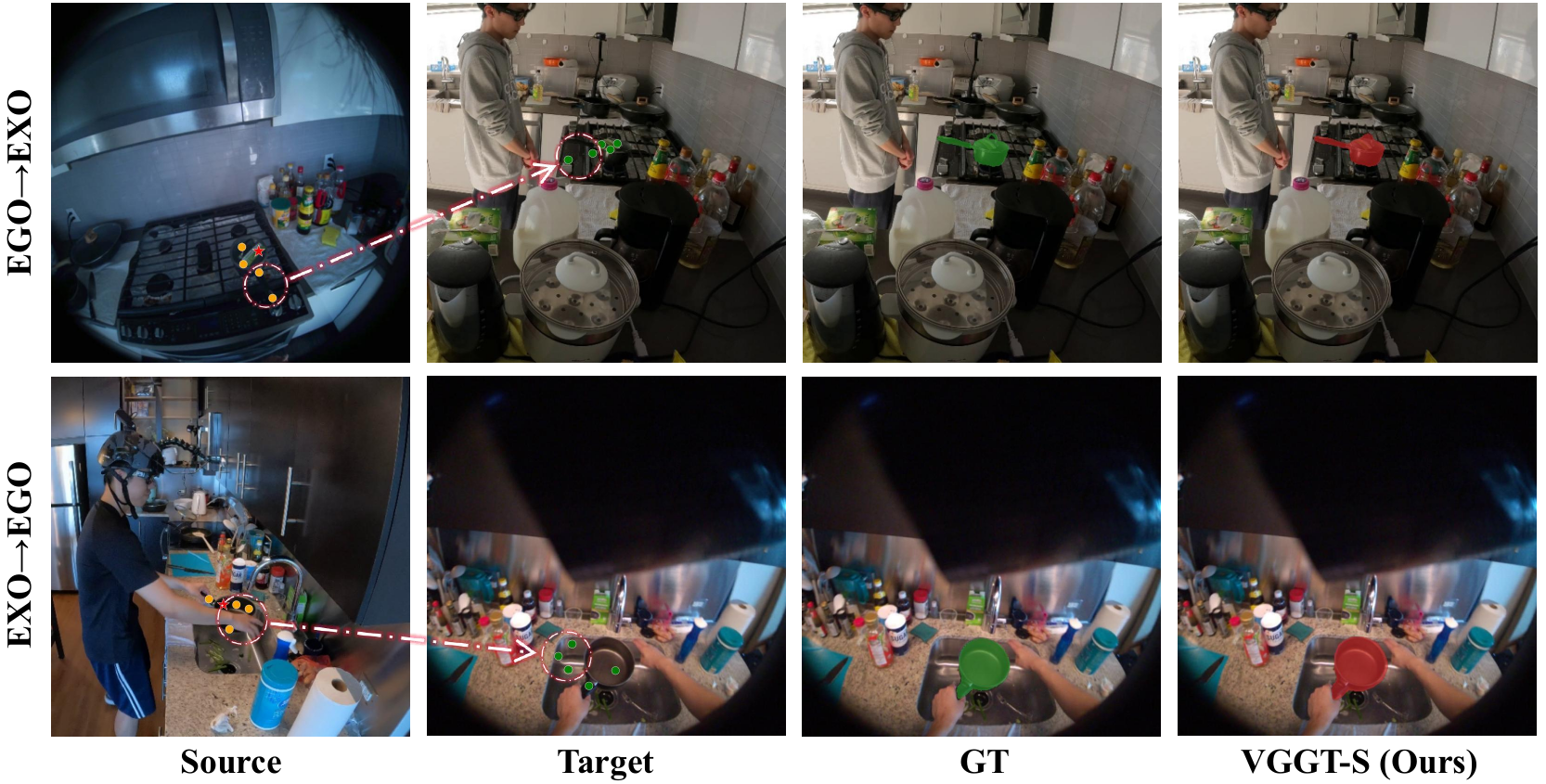} 
    \caption{\textbf{Visualization of the Effect of the Union Segmentation Head.} Although VGGT projects points to incorrect locations, our Union Segmentation Head adjusts the predicted mask to geometrically consistent positions. Zooming in provides better results.}
    \label{fig:union-head}
\end{figure}

\noindent\textbf{Visualization of the Effect of the Union Segmentation Head.}
To evaluate the effect of the Union Segmentation Head, we visualize predictions in Figure~\ref{fig:union-head}. The Union Segmentation Head explicitly aggregates contextual information while addressing the VGGT point projection bias. When raw VGGT point reprojections experience slight drift or local misalignment, the Union Segmentation Head corrects these inconsistencies through feature fusion and spatial consensus, pulling masks back to geometrically consistent locations. This results in improved alignment with the scene structure.

\noindent\textbf{Test on Outdoor Datasets.} We further assess the generalization of our correspondence-free pretrained VGGT-S on MAVREC dataset~\cite{dutta2024MAVREC}. Details and visualization can be found in the Supplementary Material.

\section{Conclusion}
We introduced VGGT-Segmentor (VGGT-S), a geometry-enhanced framework for cross-view instance-level segmentation between egocentric and exocentric perspectives. By leveraging VGGT’s geometry-consistent representations and incorporating a Union Segmentation Head with Mask Prompt Fusion, Point-Guided Prediction, and Mask Refinement, our method effectively transfers object masks across large viewpoint and scale variations. Additionally, the proposed Single-Image Self-Supervised Training strategy enables training without paired annotations, supporting Ego–Exo transfer without correspondence supervision. Extensive experiments on the Ego–Exo4D benchmark demonstrate that VGGT-S achieves state-of-the-art performance, strong generalization, offering a simple yet scalable solution for cross-view object segmentation.

{
    \small
    \bibliographystyle{ieeenat_fullname}
    \bibliography{main}
}

\end{document}